\documentclass[10pt,twocolumn,letterpaper]{article}

\usepackage[pagenumbers]{cvpr} 
\usepackage{cite}
\usepackage{amsmath,amssymb,amsfonts}
\usepackage{algorithmic}
\usepackage{textcomp}
\usepackage{graphicx,stfloats}
\usepackage{multicol}
\usepackage{multirow}	
\usepackage{booktabs}
\usepackage{cuted}
\usepackage{capt-of}
\usepackage{amssymb} 
\usepackage[T1]{fontenc} 
\usepackage{caption}
\usepackage{pythontex}
\usepackage{subcaption}
\usepackage{xcolor}

\definecolor{maroon}{cmyk}{0, 0.87, 0.68, 0.32}
\definecolor{halfgray}{gray}{0.55}
\definecolor{ipython_frame}{RGB}{207, 207, 207}
\definecolor{ipython_bg}{RGB}{247, 247, 247}
\definecolor{ipython_red}{RGB}{186, 33, 33}
\definecolor{ipython_green}{RGB}{0, 128, 0}
\definecolor{ipython_cyan}{RGB}{64, 128, 128}
\definecolor{ipython_purple}{RGB}{170, 34, 255}

\usepackage{listings}
\lstdefinelanguage{iPython}{
    morekeywords={access,and,break,class,continue,def,del,elif,else,except,exec,finally,for,from,global,if,import,in,is,lambda,not,or,pass,print,raise,return,try,while},%
    morekeywords=[2]{abs,all,any,basestring,bin,bool,bytearray,callable,chr,classmethod,cmp,compile,complex,delattr,dict,dir,divmod,enumerate,eval,execfile,file,filter,float,format,frozenset,getattr,globals,hasattr,hash,help,hex,id,input,int,isinstance,issubclass,iter,len,list,locals,long,map,max,memoryview,min,next,object,oct,open,ord,pow,property,range,raw_input,reduce,reload,repr,reversed,round,set,setattr,slice,sorted,staticmethod,str,sum,super,tuple,type,unichr,unicode,vars,xrange,zip,apply,buffer,coerce,intern},%
    sensitive=true,%
    morecomment=[l]\#,%
    morestring=[b]',%
    morestring=[b]",%
    morestring=[s]{'''}{'''},
    morestring=[s]{"""}{"""},
    morestring=[s]{r'}{'},
    morestring=[s]{r"}{"},%
    morestring=[s]{r'''}{'''},%
    morestring=[s]{r"""}{"""},%
    morestring=[s]{u'}{'},
    morestring=[s]{u"}{"},%
    morestring=[s]{u'''}{'''},%
    morestring=[s]{u"""}{"""},%
    literate=
    {á}{{\'a}}1 {é}{{\'e}}1 {í}{{\'i}}1 {ó}{{\'o}}1 {ú}{{\'u}}1
    {Á}{{\'A}}1 {É}{{\'E}}1 {Í}{{\'I}}1 {Ó}{{\'O}}1 {Ú}{{\'U}}1
    {à}{{\`a}}1 {è}{{\`e}}1 {ì}{{\`i}}1 {ò}{{\`o}}1 {ù}{{\`u}}1
    {À}{{\`A}}1 {È}{{\'E}}1 {Ì}{{\`I}}1 {Ò}{{\`O}}1 {Ù}{{\`U}}1
    {ä}{{\"a}}1 {ë}{{\"e}}1 {ï}{{\"i}}1 {ö}{{\"o}}1 {ü}{{\"u}}1
    {Ä}{{\"A}}1 {Ë}{{\"E}}1 {Ï}{{\"I}}1 {Ö}{{\"O}}1 {Ü}{{\"U}}1
    {â}{{\^a}}1 {ê}{{\^e}}1 {î}{{\^i}}1 {ô}{{\^o}}1 {û}{{\^u}}1
    {Â}{{\^A}}1 {Ê}{{\^E}}1 {Î}{{\^I}}1 {Ô}{{\^O}}1 {Û}{{\^U}}1
    {œ}{{\oe}}1 {Œ}{{\OE}}1 {æ}{{\ae}}1 {Æ}{{\AE}}1 {ß}{{\ss}}1
    {ç}{{\c c}}1 {Ç}{{\c C}}1 {ø}{{\o}}1 {å}{{\r a}}1 {Å}{{\r A}}1
    {€}{{\EUR}}1 {£}{{\pounds}}1
    {^}{{{\color{ipython_purple}\^{}}}}1
    {=}{{{\color{ipython_purple}=}}}1
    {+}{{{\color{ipython_purple}+}}}1
    {*}{{{\color{ipython_purple}$^\ast$}}}1
    {/}{{{\color{ipython_purple}/}}}1
    {+=}{{{+=}}}1
    {-=}{{{-=}}}1
    {*=}{{{$^\ast$=}}}1
    {/=}{{{/=}}}1,
    literate=
    *{-}{{{\color{ipython_purple}-}}}1
     {?}{{{\color{ipython_purple}?}}}1,
    identifierstyle=\color{black}\ttfamily,
    commentstyle=\color{ipython_cyan}\ttfamily,
    stringstyle=\color{ipython_red}\ttfamily,
    keepspaces=true,
    showspaces=false,
    showstringspaces=false,
    rulecolor=\color{ipython_frame},
    frame=single,
    frameround={t}{t}{t}{t},
    framexleftmargin=6mm,
    numbers=left,
    numberstyle=\tiny\color{halfgray},
    backgroundcolor=\color{ipython_bg},
    basicstyle=\scriptsize,
    keywordstyle=\color{ipython_green}\ttfamily,
}

\definecolor{ForestGreen}{RGB}{34,139,34}

\def\BibTeX{{\rm B\kern-.05em{\sc i\kern-.025em b}\kern-.08em
    T\kern-.1667em\lower.7ex\hbox{E}\kern-.125emX}}

\definecolor{cvprblue}{rgb}{0.21,0.49,0.74}
\usepackage[pagebackref,breaklinks,colorlinks,citecolor=cvprblue]{hyperref}

\def\paperID{16818} 
\def\confName{CVPR}
\def\confYear{2024}

\title{LEAP: LLM-Generation of Egocentric Action Programs}

\author{Eadom Dessalene, Michael Maynord, Cornelia Fermüller, Yiannis Aloimonos\\
University of Maryland, College Park\\
{\tt\small \{edessale,maynord@umd.edu\} fer@cfar.umd.edu jyaloimo@umd.edu}
}

\begin{document}
\maketitle

\begin{abstract}





We introduce LEAP (illustrated in Figure \ref{fig:programs}), a novel method for generating video-grounded action programs through use of a Large Language Model (LLM). These action programs represent the motoric, perceptual, and structural aspects of action, and consist of sub-actions, pre- and post-conditions, and control flows. LEAP's action programs are centered on egocentric video and employ recent developments in LLMs both as a source for program knowledge and as an aggregator and assessor of multimodal video information. We apply LEAP over a majority (87\%) of the training set of the EPIC Kitchens dataset, and release the resulting action programs as a publicly available dataset \href{https://drive.google.com/drive/folders/1Cpkw_TI1IIxXdzor0pOXG3rWJWuKU5Ex?usp=drive_link}{here}. We employ LEAP as a secondary source of supervision, using its action programs in a loss term applied to action recognition and anticipation networks. We demonstrate sizable improvements in performance in both tasks due to training with the LEAP dataset. Our method achieves 1st place on the EPIC Kitchens Action Recognition leaderboard as of November 17 among the networks restricted to RGB-input (see Supplementary Materials).

\end{abstract}



\section{Introduction}

\begin{figure}[h!]
    \includegraphics[width=0.5\textwidth]{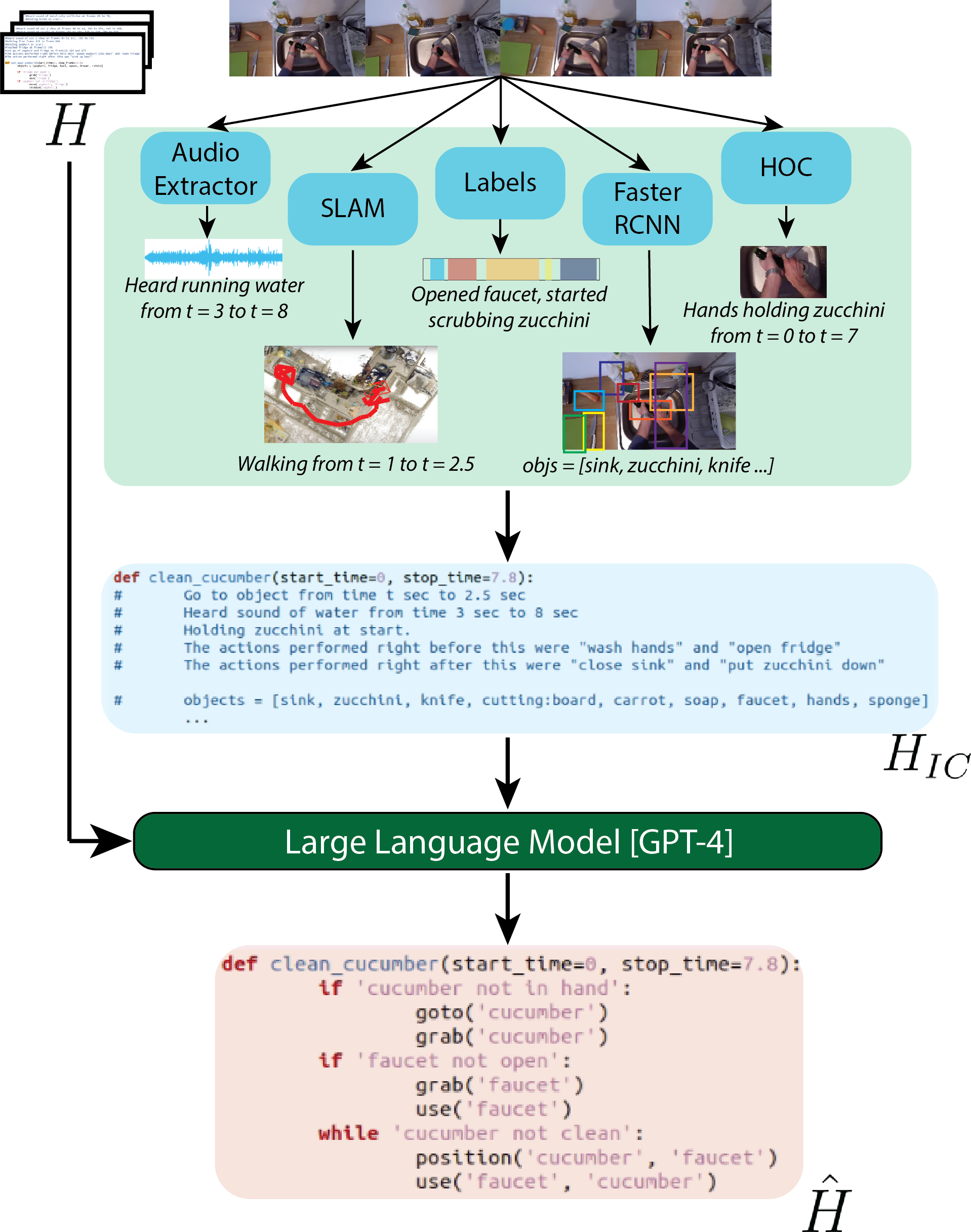}
    \caption{Illustration of the generation of egocentric action programs in LEAP. Two inputs are fed to an LLM: a pre-compiled action program library ($H$), and a textual representation of an input clip ($H_{IC}$). From this the LLM generates action program $\hat{H}$ for the input clip. Because the LLM does not take video input, $5$ components compile different aspects of the input video into text.}
    \label{fig:programs}
\end{figure}

\begin{figure*}[!t]
    \centering
    \includegraphics[width=1\textwidth]{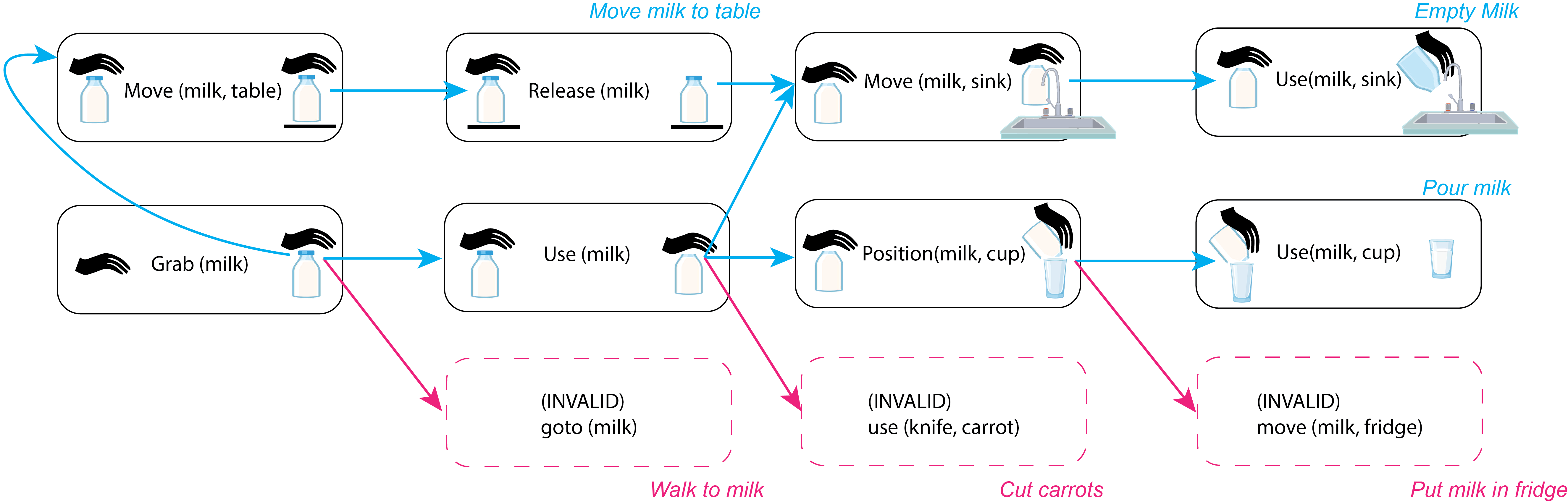}
    \caption{Illustration of chaining of primitive actions / sub-actions based on pre- and post-condition alignments.  We see that the resulting possible sequences are constrained by condition alignment (e.g. goto(milk) after grab(milk) is invalid because the milk is already in the workspace of the actor), and sequences of sub-actions group into higher level action categories (move milk to table, empty milk, pour milk). The precondition of a sub-action must match with the present world state. Valid sub-action transitions are shown with blue arrows, where red arrows are invalid sub-action transitions involve sub-action whose pre-conditions are counterfactual w.r.t. to the world state. The control flow dictates the chaining of sub-action sequences and the repetitious manner in which sub-action are applied until some condition is met, resulting in sub-action termination (e.g. use (milk, cup), use(milk, sink)).}
    \label{fig:states}
\end{figure*}

There is long thought to exist an underlying structure used in the generation of complex actions akin to natural language syntax operating over tokens of language (words) \cite{pastra2012minimalist}. One popular interpretation of this is a structuring that is hierarchical and compositional in nature, in that action can be represented as a series of steps, where each step either yields a sub-routine consisting of a series of sub-actions or yields a single terminal atomic sub-action. Existing work proposing hierarchical representations of action addresses some of the requirements for capturing this underlying structure (e.g., the ordering of multiple sub-actions into an executable sequence).




Representing actions as programs opens up an array of benefits for the interpretation and generation of complex action: 1) Action programs themselves offer \textit{robustness}, \textit{explainability}, and allow for \textit{efficiency analysis}. 2) Action programs offer an intermediary language of action between robots and humans, allowing for robots to learn from human demonstration videos by extracting action program routines which are in turn compiled and made executable by a robot. 3) Action programs are conducive to action planning and the anticipation of future action. This is because representations of action at the sensory level are less conducive than those of a higher semantic level for the anticipation of future action, motivating the need for some tokenization of behavior. Additionally, the program representation allows for the synthesis of novel action programs from a library of existing action programs. 4) Action programs also may contain control operations, which operate above the level of primitive sub-action, and below the level of action - e.g., an iterative loop with a terminating condition. 5) Action programs enable powerful \textit{zero-shot generalization}. 6) Action programs contain decompositions of unseen actions into a rich structure of seen and unseen sub-actions and conditions. Regarding points 5 and 6, the structure among these sub-actions and conditions constrain their possible interpretation, and consequently improve recognition of both the seen and unseen aspects of the action programs, leading to improved action understanding.

We propose that such a structure be represented as a \textit{program} with the complexity to model both the sub-action sequence and control flow. By grounding these action programs in egocentric video, our action programs capture properties which are visual in nature, and capture the temporal semantic character of action. This stands in contrast to existing works which often model action programs as ungrounded with respect to visual percepts \cite{singh2023progprompt,liang2023code}. In our programs, percepts are mapped onto: action primitives, pre-conditions, and post-conditions.




Action primitives are the tokenized units of sub-actions. For our primitives of choice, we include Therbligs \cite{dessalene2023therbligs}, a consistent, expressive, symbolic representation of sub-action centered around contact. We adopt $6$ Therbligs pertaining to those involving the manipulation of objects - grasp, release, move, use, position, and wait. Additionally, we include $2$ sub-actions - goto and do nothing, for a total of $8$ sub-actions.


Pre-conditions define the "why" of the sub-actions they encompass in the sense that had the pre-condition not been met, then the action would not have been executed. This stands in contrast to representations where action is represented as a sequence without pre-conditions (hence, lacking "why" to sub-action) \cite{dessalene2023therbligs}. Pre-conditions are visuo-semantic representations that capture the feasibility of a given action, and are derived from perception. Action post-conditions are the resultant outcomes of a successfully executed action. See Figure \ref{fig:states} for illustration. Nodes are centered around sub-actions which map pre-conditions onto post-conditions within each node. The precondition of a sub-action must match with the present world state. Sub-actions with pre-conditions based on counterfactual world states are invalid, and cannot be executed. From Figure \ref{fig:states}, after opening the milk the sub-action \textit{use (knife, carrot)} cannot take place as there is neither a knife or carrot held in the hand.






A program further involves a control flow dictating the order of execution of steps, whether certain sub-actions are executed in serial or parallel order, and are to be terminated or executed in repetitious manner. For the purposes of this work we focus on sub-actions executed serially, though aim to explore the parallelizability of sub-action in future robotics settings.

We exploit recent developments among Large Language Models \cite{wei2022emergent} (LLMs), using them to produce a dataset of action programs over the EPIC Kitchens dataset \cite{damen2018scaling}. The EPIC Kitchens dataset is an action dataset with hands and objects of interaction in first person view, along with narrations. 


LLMs typically do not accommodate video input - this includes the LLM that we elect to use in this work - GPT-4 \cite{OpenAI_GPT4_2023}. Figure \ref{fig:programs} illustrates the various modalities, processed independently through different sub-components. The output of each sub-component is converted into textual descriptors and subsequently provided to GPT-4. These sub-components are described in detail in Section \ref{methods}, and include:



\begin{itemize}
  \item \textbf{Audio Extractor:} We rely on pre-trained models \cite{gong2022ssast} trained over EPIC Sounds \cite{huh2023epic}, an augmentation of the EPIC Kitchens dataset with labels over audio segments.
    \item \textbf{SLAM:} We rely on EPIC Fields \cite{tschernezki2023epic}, an augmentation of the EPIC Kitchens dataset with 3D camera information based on complete reconstructions of each scene. 
    \item \textbf{Narrations:} We rely on narrations provided by the actor during, before, and after the execution of the action.
  \item \textbf{Faster-RCNN:} We rely on an open-sourced Faster-RCNN \cite{girshick2015fast} object detector provided by \cite{furnari2020rolling}. We extract all noun-level categories of all objects perceived throughout the action.
  \item \textbf{Hand-Object-Contact Detector:} We rely on the pre-trained models made publicly available at \cite{shan2020understanding}, extracting objects contacted throughout the action, which we corroborate with the Faster-RCNN to retrieve noun-level categories of contacted objects.
\end{itemize}






For each input clip in EPIC Kitchens we extract these textual descriptors, feeding them to an LLM along with hand-written exemplar action programs capturing the program specifications desired. The task of the LLM is to corroborate the different input modalities, infusing top-down knowledge into the bottom-up derived textual descriptors to produce a complete action program.

We compile a dataset of LEAP action programs covering $87\%$ of the training set of EPIC Kitchens, and make this dataset publicly available \href{https://drive.google.com/drive/folders/1Cpkw_TI1IIxXdzor0pOXG3rWJWuKU5Ex?usp=drive_link}{here}.








We employ a strong video network UniFormerV2 \cite{li2022uniformerv2}, training over our dataset of action programs for the tasks of action recognition and action anticipation. We demonstrate sizable improvements to both action recognition and anticipation due to inclusion of LEAP action programs, and among models relying solely on the RGB modality are first place on the EPIC Kitchens Action Recognition leaderboard, as of November 2023.





To recap, the primary contributions of LEAP are:
\begin{itemize}
    \item A novel \textit{formulation} of action programs - a representation flexibly integrating sub-actions, pre- and post-conditions, and which captures visual, motoric, and structural properties of action.
    \item A holistic, multi-modal approach to employing LLMs as a source of knowledge in \textit{generating} action programs.
    \item A new state-of-the-art, claiming 1st place on the EPIC Kitchens Action Recognition leaderboard (among submissions restricted to RGB input, as of November 2023). 
    \item Dataset: We release the first action program annotations over EPIC Kitchens, covering $87\%$ of the training set, available \href{https://drive.google.com/drive/folders/1Cpkw_TI1IIxXdzor0pOXG3rWJWuKU5Ex?usp=drive_link}{here}.

\end{itemize}

The rest of this paper is structured as follows: Section \ref{related} discusses Related Works, Section \ref{methods} introduces our proposed method, Section \ref{experiments} describes the experiments, and in Section \ref{conclusion} we conclude.




\section{Related Works}
\label{related}



\subsection{Structured Action Representations in Computer Vision}
There exist several video datasets that provide sub-action level annotations, enabling the compositional and hierarchical modeling of action \cite{ji2020action,shao2020finegym,shao2020intra,doughty2022you}. These datasets are typically hand-annotated, whereas the action programs generated by LEAP are done in an automatic manner (limited manual annotation necessary). Furthermore, these datasets are limited to narrow domains or are of an instructional nature \cite{xu2022finediving, qian2022svip, li2022bridge}. More recently, Therbligs \cite{salvendy2004classification,dessalene2023therbligs,dessalene2023motor} have been introduced as a low-level, mutually exclusive, contact-demarcated set of sub-actions that are flexible in application to a wide variety of datasets within
the realm of object manipulation without relying on domain
expertise. We adopt a subset of Therbligs in our action primitives of choice. 

\subsection{Large Language Models for Action Understanding}
LLMs have unified a variety of vision-language tasks \cite{lu202012, li2020unicoder, li2022blip}, with recent works applying LLMs to video \cite{momeni2023verbs,buch2022revisiting,chen2023videollm}. LLMs have been applied in obtaining dense language-level supervision for training action understanding models \cite{zhao2023learning,momeni2023verbs,zellers2022merlot}. This is found to be useful in settings where action labels are sparse, noisy, or of insufficient scale in size. With regards to the application of LLMs to action understanding, LEAP's action programs capture information otherwise missing from autogenerated captions, as these captions have a tendency to not include details of action (e.g. sub-actions, the structure of execution) \cite{park2022exposing,buch2022revisiting}.


Inspired by recent developments in LLMs in task planning \cite{singh2023progprompt,liang2023code,brohan2023can}, we adopt the task program formulation of ProgPrompt \cite{singh2023progprompt}, where a prompting scheme allows for the generation of entire executable action programs. LEAP goes beyond ProgPrompt in that while ProgPrompt derives programs solely from language, LEAP processes multi-modal information contained within video through use of an LLM. Furthermore LEAP goes beyond ProgPrompt in that LEAP captures generic egocentric activity, as opposed to ProgPrompt which is limited in application to activities within the VirtualHomes environment.





\section{Methods}
\label{methods}

\begin{figure}[ht!]
    \includegraphics[width=.47\textwidth]{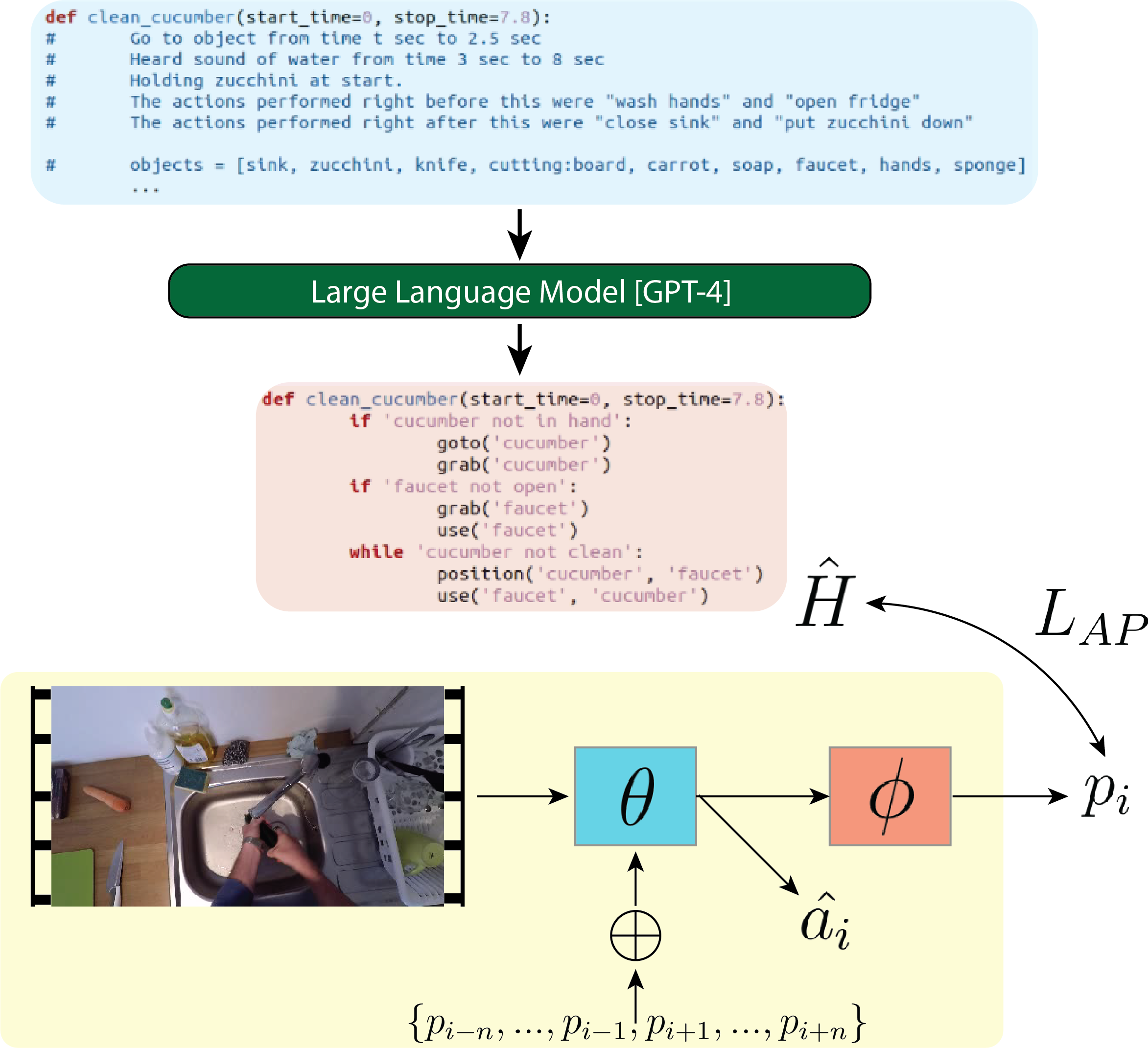}
    \caption{The relation between LEAP generated action programs, and training for action recognition and anticipation. A base action understanding network $\theta$ which predicts action $\hat{a_i}$ for clip $s_i$ is augmented with a sub-network $\phi$ to predict program $p_i$. This sub-network is trained using a loss to align it with $\hat{H}$, the LEAP action program generated for $s_i$.}
    \label{fig:architecture_combined}
\end{figure}

Given a video segment $s_i$ from EPIC Kitchens, our goal is to infer the relevant action class likelihood $\hat{a_i}$ for the tasks of action recognition and action anticipation. See Figure \ref{fig:architecture_combined} for an illustration of our complete architecture. The action program $p_i$ amounts to a compositional and hierarchical representation of $a_i$: we introduce the formulation of LEAP's action programs, discuss their generation using an LLM, and describe the dataset in Section \ref{sec:leap_dataset}. We then train a model for action understanding over this auxiliary dataset, the details of which we describe below in Section \ref{sec:learning_framework}.




\subsection{LEAP Action Programs and Dataset}
\label{sec:leap_dataset}

For video segment $s_i$, the program representation $p_i$ is a structure of discrete tokens. Each token takes the form either of a conditional statement $(while, \langle condition \rangle)$ or $(if, \langle condition \rangle )$, or of a sub-action statement $(v, o)$ nested underneath a conditional statement where $v \in V = \{\emptyset, Grasp, Release, Move, Use, Position, Goto, Wait\}$ and $o$ corresponds to the object(s) of interaction. 

Output action program $\hat{H}$ in Figure \ref{fig:programs} illustrates the LEAP action program "clean cucumber", with conditions "in hand" and "not open", and repetition with the terminating condition "not clean". The program further consists of the following sub-action sequence: "goto cucumber", "grab cucumber", "grab faucet", use("faucet"), etc.




As outlined in Figure \ref{fig:programs}, we feed two inputs into GPT4 (available through the public OpenAI API \cite{brown2020language}): $H$, and $H_{IC}$. GPT4 produces $\hat{H}$ as output. Input $H$ is a library of $20$ handwritten action programs, which include capabilities of programming languages such as import statements and comments (see Supplementary Materials). As GPT4 does not take video as input $H_{IC}$ is a textual representation, extracted through various modules, of the specific clip $s_i$. The LEAP action program $p_i$ is produced by parsing generated programs $\hat{H}$ for clip $s_i$.






Independently for each $s_i$, the program representation $p_i$ is derived from each of the following components (followed by example textual descriptors generated by said components):


  \noindent \textbf{Audio Extractor}: Describes audio of object interaction. We employ the pre-trained Self-Supervised Audio Spectrogram Transformer (SSAST) model \cite{gong2022ssast}. We first run a low-pass filter over the audio input, largely removing the background noise and extracting candidate events for classification. We then run the SSAST model in a sliding window fashion over contiguous audio, saving the classification results along with the times in which the events occurred. 
  


 \begin{lstlisting}[language=iPython]
def wipe_spoon(start_t=0, stop_t=1.63):
    # Heard sound of scrub at time
    # 0.55 sec to 0.93 sec
\end{lstlisting}

 
  \noindent \textbf{SLAM:} EPIC Fields \cite{tschernezki2023epic} provides full frame-rate 3D camera pose information using a structure from motion pipeline. The post-reconstruction fails for $29$ videos in which roughly $3\%$ of the actions are performed - we do not collect action programs over these videos. We discretize the continuous head movements, producing an output that grounds the walking behavior of the actor in time.

 \begin{lstlisting}[language=iPython]
def take_carrots(start_t=0, stop_t=4.88):
    # Go to object from time 2.57 sec to
    # time 3.77 sec 
\end{lstlisting}

 
   \noindent \textbf{Narrations:} We rely on narrations provided by the actor during, before, and after the execution of the action. These narrations provide longer-term context between actions and activities.
 
 \begin{lstlisting}[language=iPython]
def take_carrots(start_t=0, stop_t=4.88):
    # The actions performed right before
    # this were "take peppers" and
    # "take potatoes". The actions
    # performed right after this were
    # "move washing liquid" and
    # "take gravy"
\end{lstlisting}


  \noindent \textbf{Faster-RCNN:} The EPIC Kitchens 55 dataset includes bounding box annotations of objects along with noun labels. We employ the Faster-RCNN trained over this dataset, running it over each frame of $s_i$ to produce a list of objects detected. This list of objects constrains the action programs to involve only objects discovered in the scene.

 \begin{lstlisting}[language=iPython]
def carry_bowl(start_t=0, stop_t=1.93):
    # objects = [bowl, spoon, cupboard,
    # drawer]
\end{lstlisting}

 \begin{lstlisting}[language=iPython]
def align_tofu(start_t=0, stop_t=0.77):
    # objects = [tofu, knife,
    # towel_kitchen, coriander, bowl,
    # pepper]
\end{lstlisting}
  
  \noindent \textbf{Hand-Object-Contact Detector:} The pre-trained egocentric model provided by \cite{shan2020understanding}, produces contact states for each hand. We employ it over each frame of $s_i$, producing a sequence of contact states for each hand. As the model only produces masks of objects in contact, we retrieve the classes of contacted objects by computing the intersection-over-union w.r.t. the objects detected in the scene with the previous Faster-RCNN. This component serves to constrain the generation of the action programs to only objects that play a direct role in the action.

 \begin{lstlisting}[language=iPython]
def take_carrots(start_t=0, stop_t=4.88):
    # Holding nothing at start. 
    # Grabbed carrots at time(s) 2.82
\end{lstlisting}

 \begin{lstlisting}[language=iPython]
def takeonion_fridge(start_t=0, stop_t=3):
    # Released fridge at time(s) 2.75
\end{lstlisting}

Our dataset spans $58,000$ of the $67,217$ action clips contained within the training set of the EPIC Kitchens dataset. Each query of $H_{IC}$ contains $35$ programs to be completed, for a total cost per query of $\$0.17$. We refer the reader to the Supplementary Materials for more details.

\begin{figure*}[!t]
    \centering
    \includegraphics[width=1.\textwidth, height=240pt]{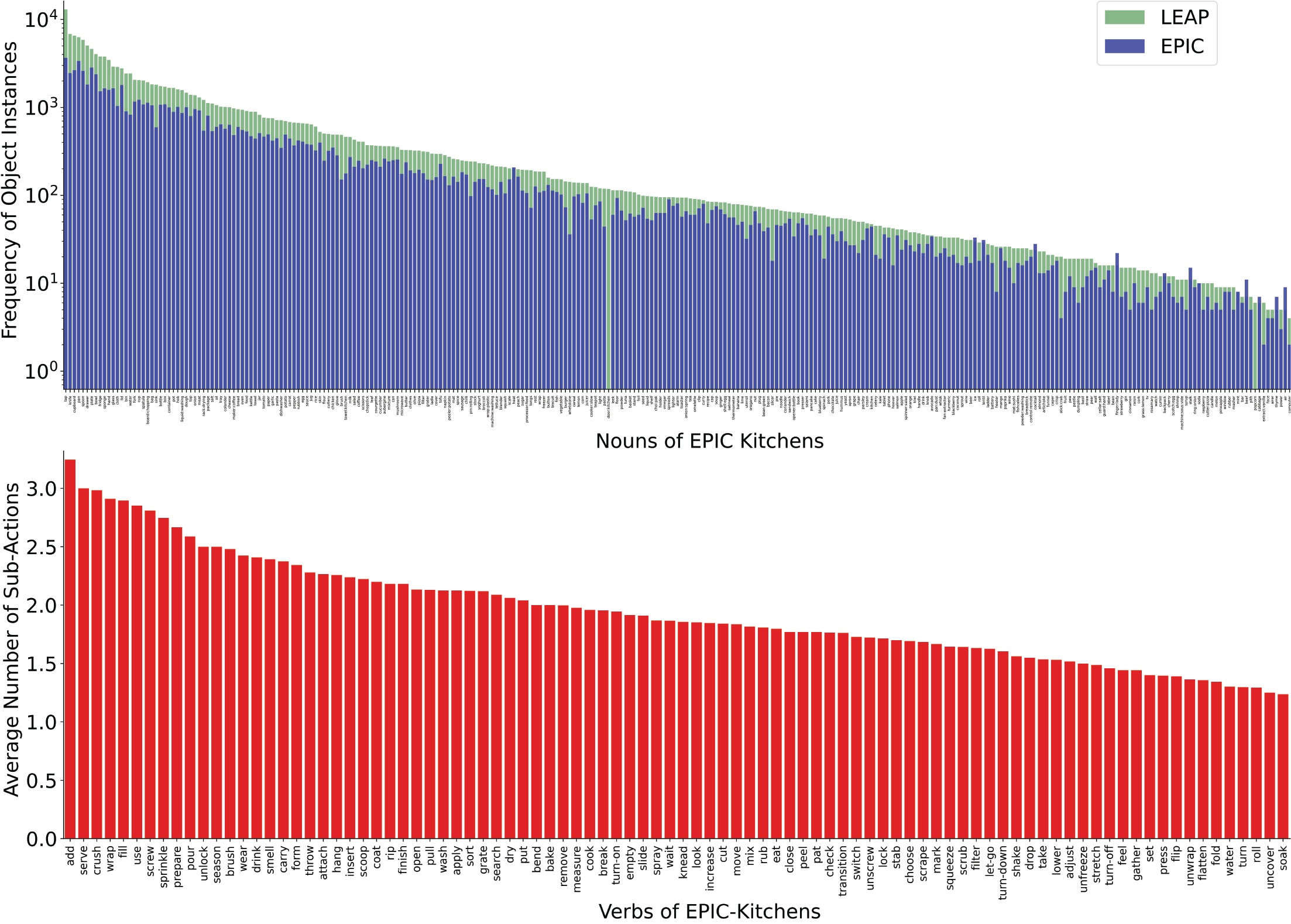}
    \caption{\textbf{(TOP)} Stacked histogram (y-axis is logarithmically scaled) of object frequency over the object categories collected in the labels of the EPIC Kitchens dataset ({\color{blue} blue}), versus the object categories collected over the action program dataset generated by LEAP ({\color{ForestGreen} green}). \textbf{(BOTTOM)} Histogram of sub-action frequency for verb categories of the EPIC Kitchens action ontology, within the action program dataset generated by LEAP.}
    \label{fig:histograms}
\end{figure*}

The lower histogram of Figure \ref{fig:histograms} illustrates the number of sub-actions contained within each action program for each of the 97 verb classes in the EPIC Kitchens dataset. Verbs expected to take longer and involve multiple steps (e.g. \textit{serve}, \textit{prepare}, \textit{wrap}) contain a higher number of sub-actions than verbs of a shorter duration involving fewer steps (e.g. turn, take, soak).




The upper histogram of Figure \ref{fig:histograms} illustrates distributions of object frequencies, across our dataset and EPIC Kitchens, for the EPIC Kitchen object ontology. EPIC Kitchens annotations exhibit a long-tailed object distribution, in part a consequence of annotators often failing to annotate the involvement of certain objects (e.g., "knife-sharpener" is often not annotated for the action "sharpen knife"). This stands in contrast to the object frequency distribution of our dataset, which does not exhibit such a long-tailed distribution. Note the y-axis is exponentially scaled - our LEAP object annotations contain multiple times the number of objects as those in EPIC Kitchens.



\subsection{Learning Framework}
\label{sec:learning_framework}

Figure \ref{fig:architecture_combined} illustrates the relation between LEAP action program generation and the training of an action network. The action network is augmented with a head to predict programs, which is trained with a loss term aligning it to the LEAP generated action program.

As our modification of a generic video understanding architecture takes place at the penultimate layers of the network, the primary base architecture for our approach can be any popular video architecture. Our base network of choice in this work is the UniFormerV2 model \cite{li2022uniformerv2}, a state-of-the-art model that unifies convolutions and self-attention within a Transformer network. We adopt the pre-trained Kinetics-400 weights provided at the model zoo of the Github repository \href{https://github.com/OpenGVLab/UniFormerV2}{here}.

\begin{figure}[ht!]
    \includegraphics[width=0.48\textwidth]{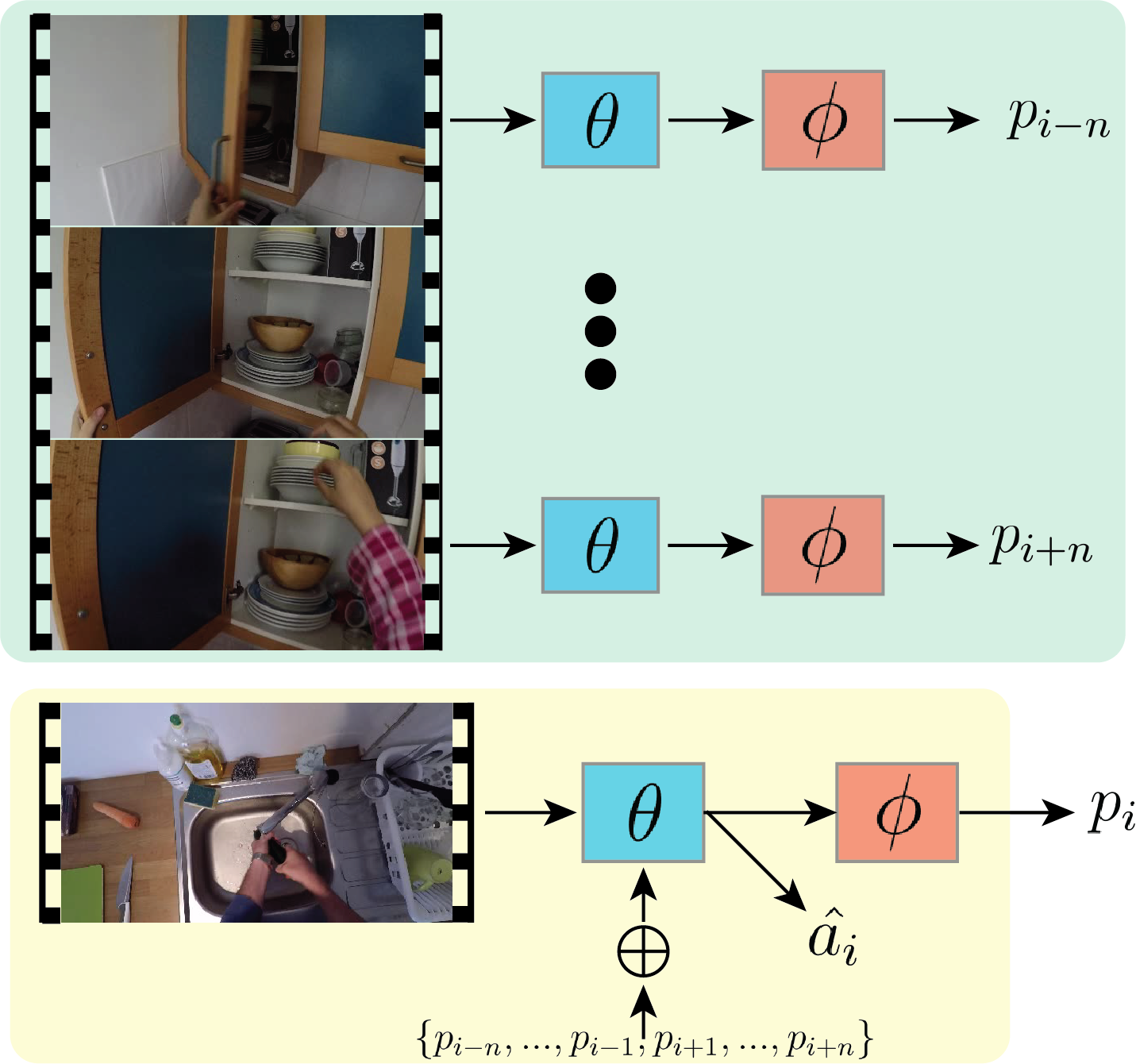}
    \caption{We independently feed input clip $s_i$ as well as surrounding video clips $s_q$, producing programs $p_q$ for $i - n \leq i \leq i + n$. We then feed all global video tokens extracted from $\theta$ for each $q$ along with global video tokens associated with video input $s_i$ to a fully connected layer ($\phi$), producing action prediction $\hat{a_i}$.}


    \label{fig:architecture_single}
\end{figure}

The layout of our video architecture is shown in Figure \ref{fig:architecture_single}. We apply two small modifications to the base network: 1) a time embedding is added to the extracted video features after the application of the 3D convolutional feature extractor, and 2) we append $\phi$, consisting of a fully connected head and a GRU module over the final global video token produced by the UniFormerV2 network - see \cite{li2022uniformerv2} for more. The fully connected head produces pre-condition predictions for input video clip $s_i$. The GRU module is followed by two fully connected heads, one for sub-action verb $v$ and another for sub-action object $o$, producing a sequence of sub-action predictions $(v, o)$ for input video clip $s_i$. 

In addition, we briefly train a second (identical) UniFormerV2 network for only $5$ epochs against action program loss (no action labels) discussed in \ref{sec:learning_framework}, freezing it during the training of the primary action classification architecture. We do this as the EPIC Kitchens dataset is not densely labeled. With this second network, we extract action programs for contextual video clips $s_q$ for $i - n \leq q \leq i + n$ subject to $q \neq i$, producing video tokens for all clips $s_q$ where $n$ is a window size that varies by task. These global video tokens $s_q$ are added to global video tokens extracted from $s_i$, over which action classification is performed. The fully connected layer in UniFormerV2 producing action class likelihoods $\hat{a_i}$ is left untouched. 


We adopt two auxiliary loss functions in producing our programs of action - $L_T$ (cross-entropy loss comparing predicted sub-actions and ground truth sub-actions) and $L_{PC}$ (L1 loss comparing predicted pre-conditions and ground truth pre-conditions) in addition to the original cross-entropy loss $L_{CE}$ over the model logits and the ground truth action labels. Our final combined loss is $L_{AP} = L_T + L_{PC} + L_{CE}$.

In deriving our ground truth pre-conditions we aggregate separate preconditions within $p_i$ into a single sentence (e.g. "if cucumber not in hand" and "if faucet not open" becomes "if cucumber not in hand and if faucet not open" over which a single sentence embedding $e_i$ using \cite{reimers-2019-sentence-bert} is computed. The $L_{PC}$ loss is computed as the L1-norm between the sentence embedding representation of the aggregated pre-condition and the regressed output of the fully connected head.


In deriving our ground sub-action sequence, we flatten all sub-actions within $p_i$ into a single sequence (regardless of pre-condition) $(v_1, o_{1_1}, o_{1_2}), (v_2, o_{2_1}, o_{2_2}), ..., (v_{n}, o_{n_1}, o_{n_2})$ where $n < 10$. Through quantitative and qualitative evaluation we find the following constraint relaxations to be beneficial:

\begin{enumerate}
    \item Rather than devote two separate object heads for sub-actions involving two objects, we devote a single fully connected head to the prediction of all sub-action objects, replacing hard object labels $o_{z_1}$ and $o_{z_2}$ for $1 \leq z \leq n$ with soft object label $o_z$.
    \item Rather than compute $L_T$ based on strict ordering between predicted sub-action sequences and ground-truth sub-action sequences, we employ a containment-based loss, where predicted sub-actions are penalized when not present within the ground-truth sub-action sequence.
\end{enumerate}

\section{Experiments}
\label{experiments}




Our ablation experiments explore the extent to which we are able to predict programs and the extent to which their incorporation benefits action recognition and action anticipation. All experiments are conducted over the EPIC Kitchens dataset. See Section \ref{prog_abl} for an ablation study over the prediction of action programs and Section \ref{ablat} for ablations over their incorporation. We describe our submission to the EPIC Kitchens leaderboard in Section \ref{comp}, the final results of which are included in the Supplementary Materials.

\subsection{Training Details}
\label{sec:training_details}

For all ablations, we adopt the $32$ frame input UniFormerV2 with an identical model architecture as that trained over Kinetics \cite{carreira2017quo} in \cite{li2022uniformerv2}. We adopt a base learning rate of $5e-6$, a batch size of $16$, and employ $4$ A100 GPUs for all training runs. As the EPIC Kitchens dataset is smaller in size than those used in \cite{li2022uniformerv2}, we adopt stronger data augmentations and use more training epochs in action recognition ($30$ epochs). We train for only $20$ epochs over the task of action anticipation. See \cite{li2022uniformerv2} for other details (under large configuration fine-tuned from K710 followed by K400).

\subsection{Action Program Recognition}
\label{prog_abl}
Our dataset of action programs spans roughly $87\%$ of the EPIC Kitchens training set - our action programs do not span the validation or test set of the EPIC Kitchens dataset. As such, we split our dataset by participant ID, reserving the first $32$ participant homes for training and the last $5$ participant homes for evaluation.

As our model's pre-condition predictions are continuous sentence-level embeddings (and not discrete tokens), we focus evaluation on the comparison between the predicted sub-action and ground truth sub-action sequences. See Table \ref{prog_abl} where S-Verb corresponds to the verb of a sub-action, Obj (A) corresponds to the first ("acting") object of interaction, and Obj (B) corresponds to the second ("recipient") object of interaction. Accuracy is based on set-level comparison between ground truth and predicted sequences, and not based on strict ordering requirements (see \ref{sec:learning_framework}).

\subsection{Action Program Ablations for Action Recognition \& Anticipation}
\label{ablat}
We evaluate the extent to which our action program formulation benefits action recognition and anticipation, with results shown in Tablea \ref{table:action_recognition} and \ref{table:action_anticipation}. We perform the following comparisons: \textbf{Base} corresponds to the large UniFormerV2 architecture in \cite{li2022uniformerv2}, \textbf{$L_{AP}$ w/ Aggr} corresponds to the UniFormerV2 architecture where we incorporate loss component $L_{AP}$ (see Section \ref{sec:learning_framework}), \textbf{Aggr} corresponds to the base architecture provided temporal context (feeding video clips $s_q$ for $i - n \leq q \leq i + n$) without training against loss component $L_{AP}$ (only training action prediction network against action labels) and \textbf{Full} corresponds to the entirety of our proposed framework, both aggregating video clips and training against action program loss $L_{AP}$.

In the action recognition setting, it is allowed to provide temporal context both before and after the duration of the action, and so we set $n$ to $2$ where each $s_q$ spans an identical number of frames as that of the input video clip $s_i$. However, in action anticipation, the model cannot observe input beyond frame $t_s - \tau_a$ where $t_s$ is the start time of the action and $\tau_a$ is the anticipation time (set to $1$ second for all experiments reported). As such, when aggregating video representations we provide $s_q$ for $i - 2 \leq q \leq i$ where each $s_q$ spans $256$ frames.

\begin{table}[t!]
\centering
\resizebox{0.46\textwidth}{!}{%
\begin{tabular}{l*{5}c}
\cmidrule{0-4}
 Losses & S-Verb & Obj (A) & Obj (B) & \\
 \cmidrule{0-4}
 $L_{T}$ & \textbf{61.56\%} & 57.27\% & 36.26\% & \\
 $L_{T}+L_{CE}$ & 60.75\% & 60.45\% & 39.43\% &  \\
 $L_{T}+L_{PC}$ & 60.57\% & 58.39\% & 37.77\% &  \\
 $L_{T}+L_{PC}+L_{CE}$ & 60.62\% & \textbf{60.82\%} & \textbf{39.93\%} &  \\
\end{tabular}}
\caption{Sub-action recognition results over our validation split. S-Verb corresponds to sub-action level verbs (as opposed to action level verbs), \textbf{Obj (A)} corresponds to recognition of active objects, \textbf{Obj (B)} corresponds to recognition of objects acted upon.}
\label{table:program_ablation}
\end{table}

\subsection{EPIC Kitchens Challenge}
\label{comp}
We compare our final architecture against architectures belonging to other approaches through comparison over the EPIC Kitchens action recognition leaderboard. We describe the Top-3 submissions posted on the leaderboard that refrain from incorporating modalities other than RGB images (e.g. optical flow or sound). We demarcate the following comparisons by team name, followed by a description of their proposed method: \textbf{SCUT - JD} \cite{jiang20221st} employs a four model ensemble of SlowFast networks trained on different alterations of the ground truth, \textbf{NUS-HUST-THU-Alibaba} \cite{sudhakaran2021saic_cambridge} employs a classic video vision transformer architecture from \cite{arnab2021vivit}, \textbf{SOS-OIC} \cite{escorcia2022sos} employs a TSM \cite{lin2019tsm} model trained using a novel self-supervised approach. For the reporting of the EPIC Kitchens results, we report test accuracy after training for $30$ epochs ensembling only two models. As leaderboard results are ongoing, we include results in the Supplementary Materials (leaderboard closes November 25).


\begin{table}[t!]
\centering
\resizebox{0.46\textwidth}{!}{%
\begin{tabular}{l*{5}c}
\cmidrule{0-4}
 Ablations & Verb & Object & Action & \\
 \cmidrule{0-4}
 $Base$ (w/ Aggr) & 68.54\% & 59.91\% & 48.19\% & \\
 $L_{AP}$ (w/ Aggr) & 69.95\% & 59.85\% & 49.10\% &  \\
 Aggr (w/ $L_{AP}$) & 68.40\% & 60.23\% & 48.31\% &  \\
 Full & \textbf{70.09\%} & \textbf{61.03\%} & \textbf{50.26\%} &  \\
\end{tabular}
}
\caption{Action recognition accuracies over EPIC Kitchens dataset. Results under EPIC Kitchens are provided as:
verb/object/action prediction accuracies,}
\label{table:action_recognition}
\end{table}

\begin{table}[t!]
\centering
\resizebox{0.46\textwidth}{!}{%
\begin{tabular}{l*{5}c}
\cmidrule{0-4}
 Ablations & Verb & Object & Action & \\
 \cmidrule{0-4}
 $Base$ (w/ Aggr) & 35.20\% & 33.92\% & 14.64\% & \\
 $L_{AP}$ (w/ Aggr) & 35.96\% & 33.89\% & 14.76\% &  \\
 Aggr (w/ $L_{AP}$) & 35.17\% & 35.78\% & 15.03\% &  \\
 Full & \textbf{38.12\%} & \textbf{36.92\%} & \textbf{16.98\%} &  \\
\end{tabular}}
\caption{Action anticipation accuracies over EPIC Kitchens. Results under EPIC Kitchens are provided as:
verb/object/action prediction accuracies,}
\label{table:action_anticipation}
\end{table}



\section{Discussion}
We observe that incorporation of action programs improves all action recognition \& anticipation results. We note that benefits to action anticipation are particularly sizable, attributable to the longer-term temporal semantics captured through the aggregation of action program predictions over time.

\begin{figure}[!t]
    \centering
    \includegraphics[width=.47\textwidth]{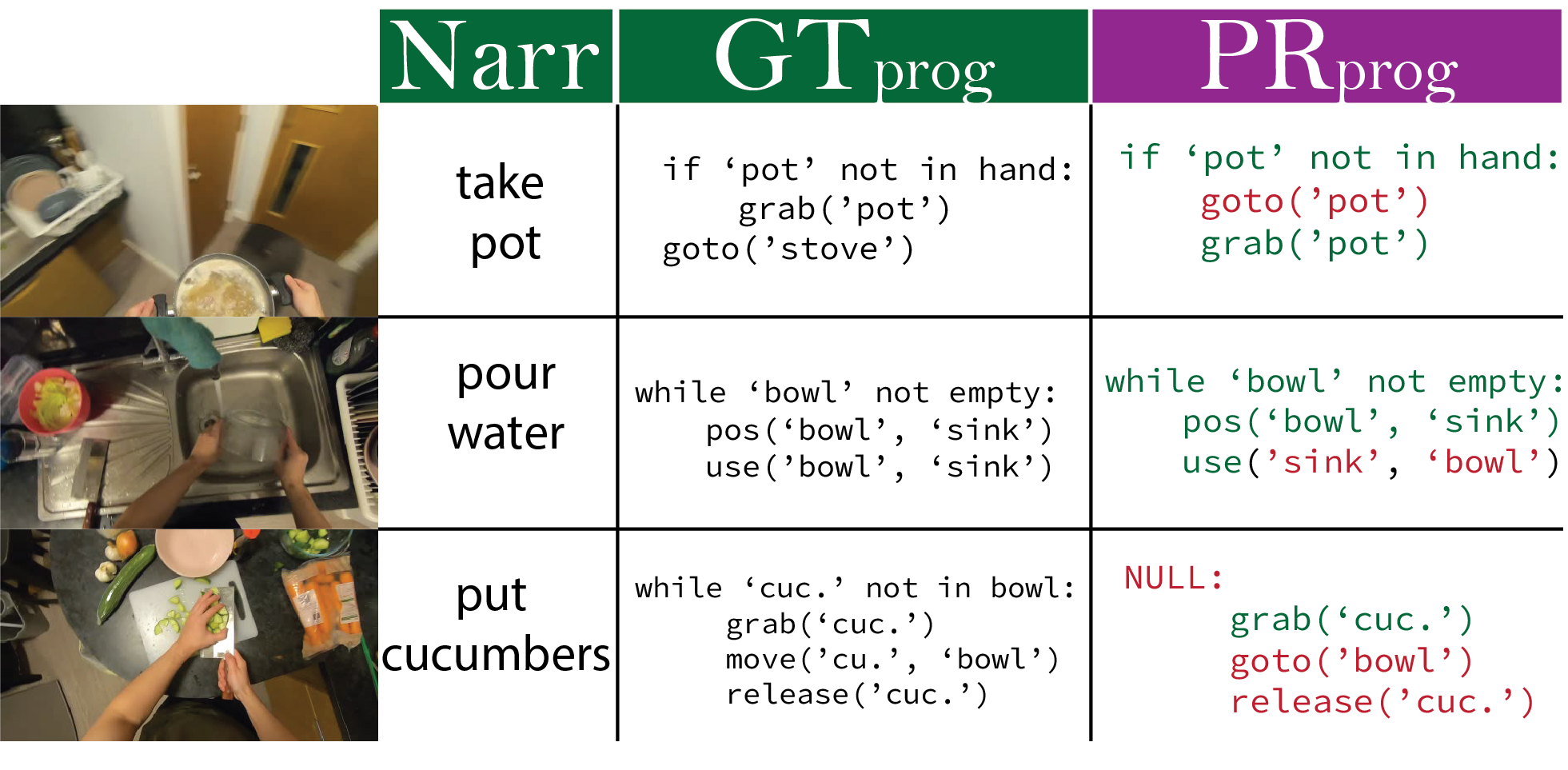}
    \caption{Qualitative assessment of the recognition of action programs from video (without the loss constraint relaxations discussed in Section \ref{sec:learning_framework} for purposes of illustration). $Narr$ corresponds to EPIC Kitchens dataset narrations, $GT_{prog}$ corresponds to the ground truth action program and $PR_{prog}$ corresponds to the predicted program. To select the pre-conditions in example $PR_{prog}$ shown we apply a threshold over distance between $GT_{prog}$ and $PR_{prog}$ when projected into an embedding space. Sub-action sequences are directly compared between $GT_{prog}$ and $PR_{prog}$. We select these instances to show current limitations of LEAP. Correct predictions are highlighted in green, incorrect predictions are highlighted in red.}
    
    \label{fig:qual}
\end{figure}

Training the action understanding network to be sensitive to action program ordering results in mode collapse, to the detriment of performance over clips with non-standard sub-action sequences. See Figure \ref{fig:qual} for failures in predicting actions performed in non-standard sequences. We find relaxing the sequence in the loss results in slightly improved downstream action understanding tasks. We also discover the network has trouble differentiating between objects "acting" and objects "being acted upon". Devoting separate object heads for disambiguation resulted in weaker object-level performance in sequence prediction and downstream action recognition tasks.





\begin{figure}[!b]
    \centering
    \includegraphics[width=.45\textwidth]{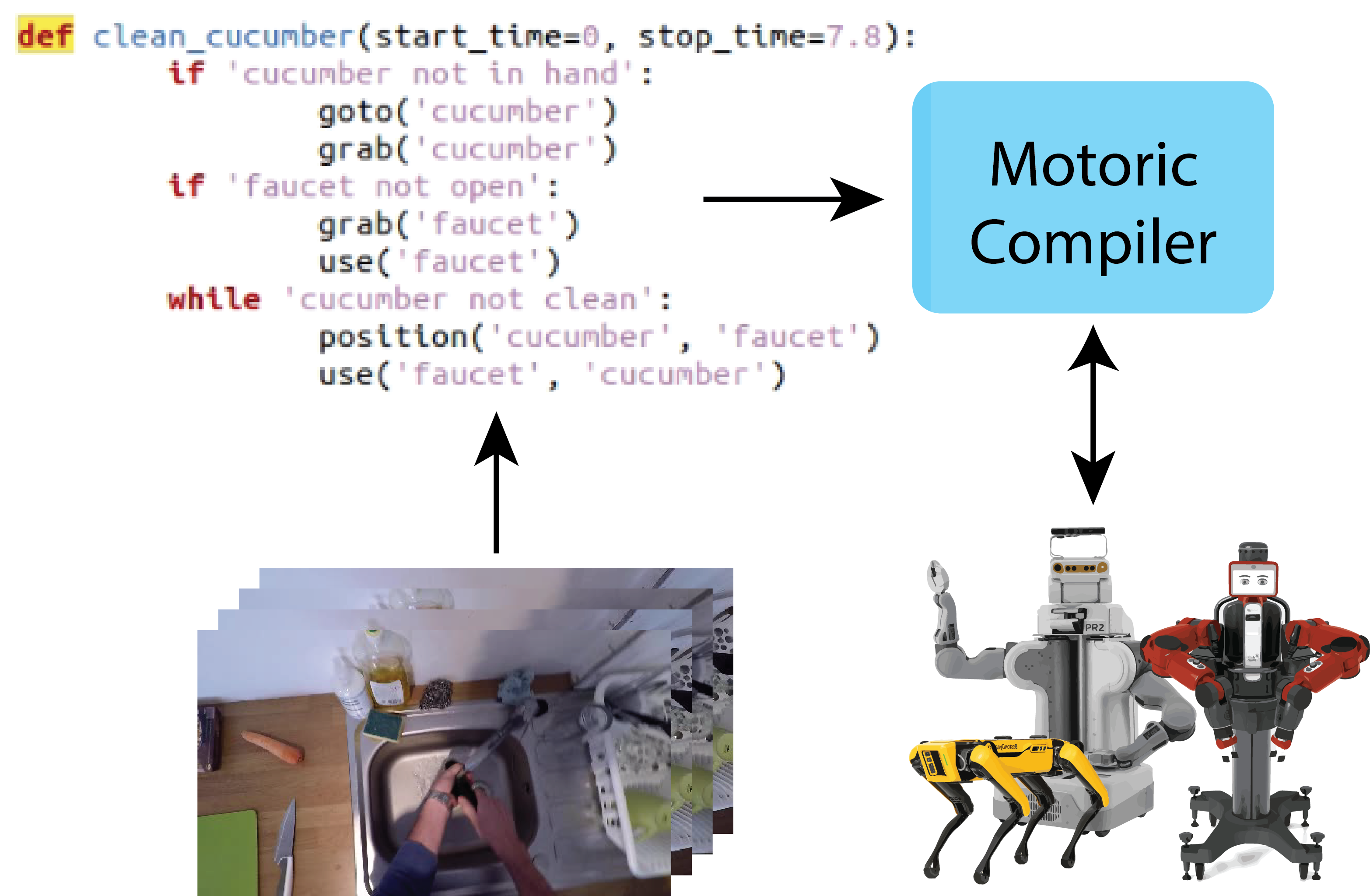}
    \caption{LEAP's action programs are rooted in visual perception, but are also actionable. Therefore LEAP can serve as a bridge between perception and action, enabling robot learning from demonstration. As the action representations in LEAP are not platform specific, LEAP action programs could be run across multiple robotic platforms.}
    \label{fig:robot}
\end{figure}

Future work includes the development of action understanding architectures with inductive biases better suited to hierarchical representations of action (as is reflected in LEAP's action programs). LEAP also facilitates additional tasks to explore: long-form video understanding to zero-shot/few-shot action recognition, etc.

As LEAP's action programs function both as a perceptual and an actionable representation they have application to cognitive theories of action, and observations such as the existence of mirror neurons which draw a connection between action perception and execution. We hope that LEAP stimulates research on the rich interplay between the interpretation of action and the generation of action (e.g., cross-platform robot learning from demonstration in Figure \ref{fig:robot}).



\section{Conclusion}
\label{conclusion}


We introduce LEAP, a novel method for generating video-grounded action programs through use of an LLM. LEAP captures the motoric, perceptual, and structural aspects of action. We apply LEAP over 87\% of the training set of EPIC Kitchens, and release the resulting action programs. We demonstrate the utility of LEAP, training a network for action recognition and anticipation over the LEAP dataset, resulting in sizable empirical benefits to action understanding from the incorporation of action programs. Our method achieves 1st place on the EPIC Kitchens Action Recognition Challenge among the networks restricted to RGB input. We are excited about exploring new architectures, action understanding tasks, and program formulations, and see this work as leading to many future possibilities.



\bibliographystyle{unsrt}

\bibliography{egbib}


\end{document}